\documentclass[letterpaper, 10 pt, conference]{ieeetran}  

\IEEEoverridecommandlockouts                              

\usepackage{graphicx} 
\usepackage{epsfig} 
\usepackage{mathptmx} 
\usepackage{times} 
\usepackage{amsmath,amssymb,amsfonts}
\usepackage{cite}
\usepackage{pgfplots}
\pgfplotsset{compat=1.7}
\usepackage{pgfplotstable}
\usepackage{tikz,pgfplots}
\usepackage{physics} 
\usepackage{siunitx} 
\usepackage{enumerate} 
\usepackage{tikz,pgfplots}
\usepackage{verbatim}
\usepackage{color}
\usepackage{mathtools}
\usepackage{breqn}
\usepackage{subcaption}
\usepackage{dblfloatfix}
\usepackage{comment}

\usepackage{tikz}

\title{PAPRAS: Plug-And-Play Robotic Arm System}

\author{Joohyung Kim, Dhruv C Mathur, Kazuki Shin, Sean Taylor
\thanks{ Joohyung Kim, Dhruv C Mathur, Kazuki Shin, and Sean Taylor are with the KIMLAB (Kinetic Intelligent Machine LAB) at the University of Illinois Urbana-Champaign
     {\tt\small{\{joohyung, dmathur2, kazukis2, seanlt2\}}@illinois.edu}. All authors contributed equally to this work (alphabetical order).}%
}

\begin{document}
\maketitle
\thispagestyle{empty}
\pagestyle{empty}

\begin{abstract}
This paper presents a novel robotic arm system, named PAPRAS (Plug-And-Play Robotic Arm System). PAPRAS consists of a portable robotic arm(s), docking mount(s), and software architecture including a control system. By analyzing the target task spaces at home, the dimensions and configuration of PAPRAS are determined. PAPRAS’s arm is light (less than 6kg) with an optimized 3D-printed structure, and it has a high payload (3kg) as a human-arm-sized manipulator. A locking mechanism is embedded in the structure for better portability and the 3D-printed docking mount can be installed easily. PAPRAS’s
software architecture is developed on an open-source framework and optimized for low-latency multiagent-based distributed manipulator control. A process to create new demonstrations is presented to show PAPRAS’s ease of use and efficiency. In the paper, simulations and hardware experiments are presented in various demonstrations, including sink-to-dishwasher manipulation, coffee making, mobile manipulation on a quadruped, and suit-up demo to validate the hardware and software design.
\end{abstract}

\section{Introduction}
Robotic manipulation has been explored since the start of robotics research. Industrial robotic arms have been developed and used for decades for fast and precise manufacturing and heavy-loaded tasks. From car assembly to semiconductor fabrication, robotic arms and manipulation have made a significant contribution to the manufacturing industry.

Recently, robotic manipulation has broadened its boundaries for various manipulation tasks, including cooperative manipulation tasks between humans and robots. Many robotic arms for pHRI (physical Human-Robot Interaction) were commercialized, such as 
the LBR iiwa from Kuka~\cite{Bischoff2010}, the robotic arms from Kinova~\cite{Kinova}, and Franka robot arms from Franka Emika~\cite{Haddadin2022}. These robotic arms have force and/or torque sensors to consider safe interaction when there are contacts or collisions with humans in the workspace. 
In addition to the force/torque sensors, mechanically compliant joints and lightweight mechanisms with high back-drivability have been investigated to ensure safe interaction. One of the most popular approaches for the compliant joint design is having an elastic component in the mechanism~\cite{Pratt1995}. The lightweight mechanism can be achieved by locating the heavy actuators near the robot base and utilizing a transmission mechanism (e.g. cable-driven mechanism) for joint actuation~\cite{Salisbury1988, Kim2017}. 

\begin{figure}
\centering
\includegraphics[width=1\columnwidth]{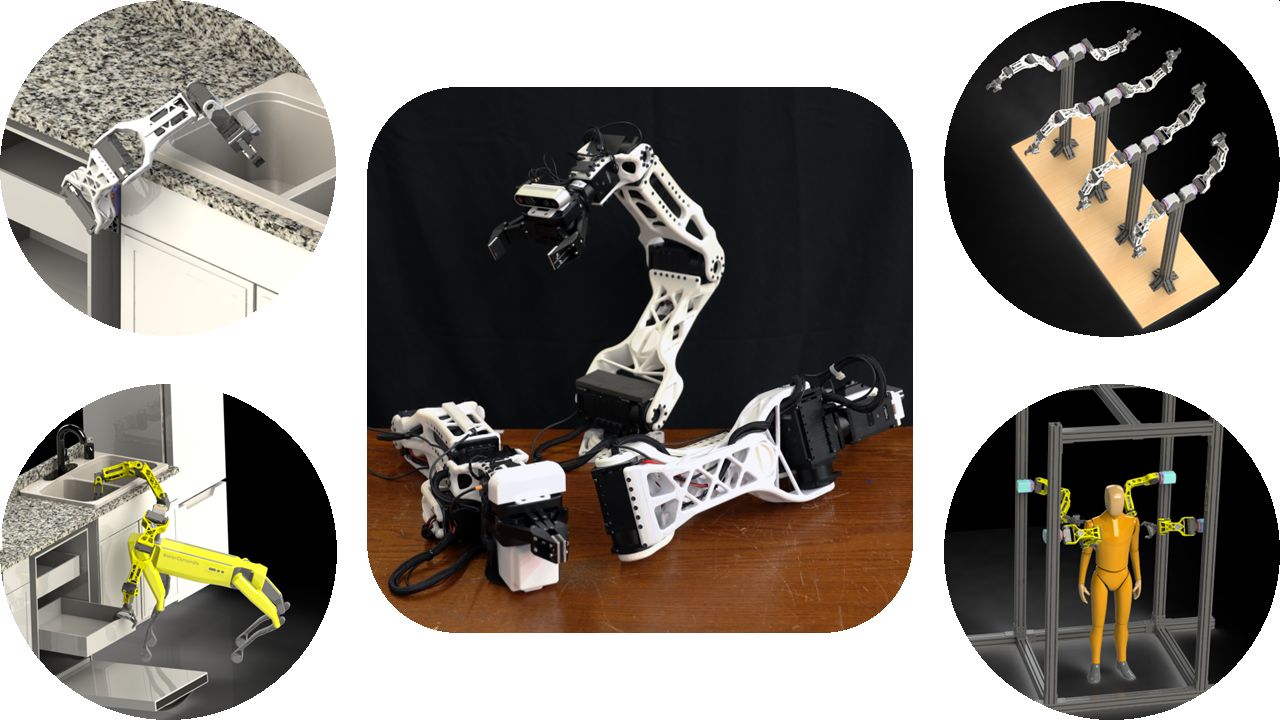}
\caption{PAPRAS mounted in various environments using the plug-and-play feature and executing manipulation tasks.}
\label{fig:papras} 
\end{figure}

Despite continued achievements and efforts, robotic manipulators are not ready for home use yet. Most existing robotic arms are not affordable (higher than \$20,000 USD) and not lightweight (more than 10kg). Additionally, they must be bolted down to a fixed structure or need mechanical clamps to hold the base of the robot, and it is hard to move once installed. Although the home environment is well-segmented depending on the purposes, such as a kitchen for cooking and a utility room for laundry, 
it is cumbersome and time-consuming if the arm needs assembly and disassembly every time when moving from one place to another. Some researchers are trying to utilize mobile manipulators~\cite{Yamamoto2019, Kemp2022}, but these are ongoing efforts to make them compact, dexterous, and safe.


In this paper, we introduce a new robotic manipulator, PAPRAS (Plug-And-Play Robotic Arm System), to tackle this challenge by designing a docking system with a modular, plug-and-play robotic arm and optimizing its design. Our primary objective was to create a pluggable, portable, and lightweight manipulator capable of performing a variety of tasks as shown in Fig.~\ref{fig:papras}. There are several research works related to pluggable robotic limbs. Topping~\cite{Topping2001FlexibotA} introduced an interesting concept of Flexibot which has the pluggable feature to a wheelchair or a wall mount, but it has not been implemented. In modular robot research, researchers have been working on modular, reconfigurable limbed robots~\cite{Kim2017Snapbot, Gim2020, Hu2022, Zhu2022}. Also, a company named GITAI recently released large-scale pluggable robotic arms for space~\cite{GITAI}. The proposed PAPRAS is designed for home and human collaboration. Within this paper, we present the design of PAPRAS and provide empirical evidence for its efficacy by showing its applications.

The paper is organized as follows. In Sections II and III, we introduce the hardware design of the proposed robotic arm. Section IV explains PAPRAS's software architecture. Validations and applications of PAPRAS are presented in Section V. Lastly, the conclusion and future work are discussed in Section VI.

\section{Mechanical Hardware Design}

\begin{figure}[b]
    \centering
    \begin{subfigure}{0.38\columnwidth}
    \centering
    \includegraphics[width = .9\columnwidth]{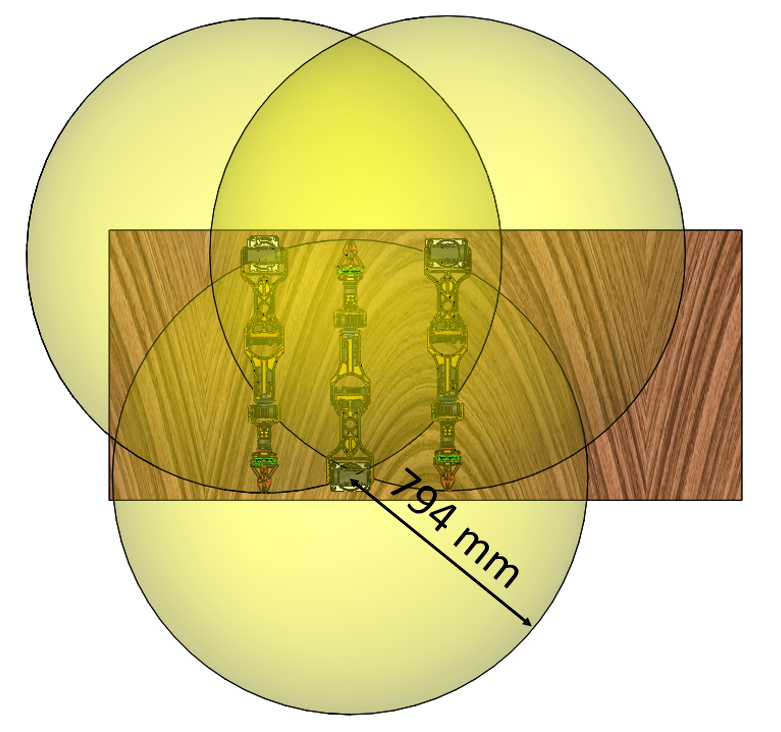}
    \caption{Table (top)}
    \label{fig:table_top}        
    \end{subfigure}
    \hfill
    \begin{subfigure}{0.29\columnwidth}
        \centering
        \includegraphics[width = .95\columnwidth]{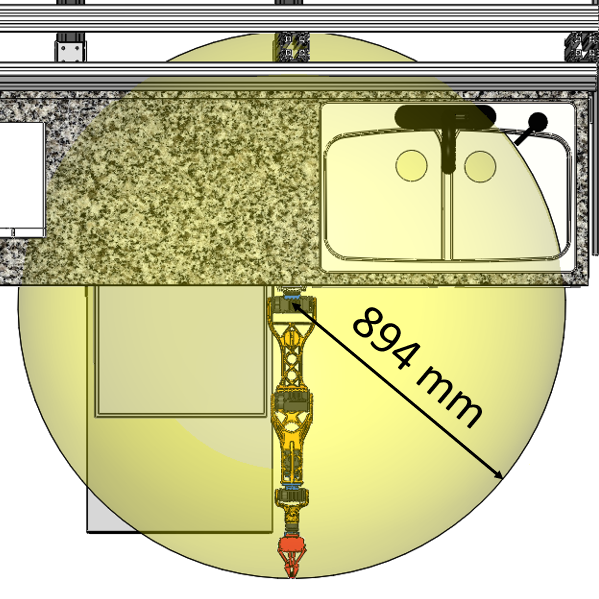}
        \caption{Kitchen (top)}
        \label{fig:kitche_up}
    \end{subfigure}
    \hfill
        \begin{subfigure}{0.29\columnwidth}
        \centering
        \includegraphics[width = .98\columnwidth]{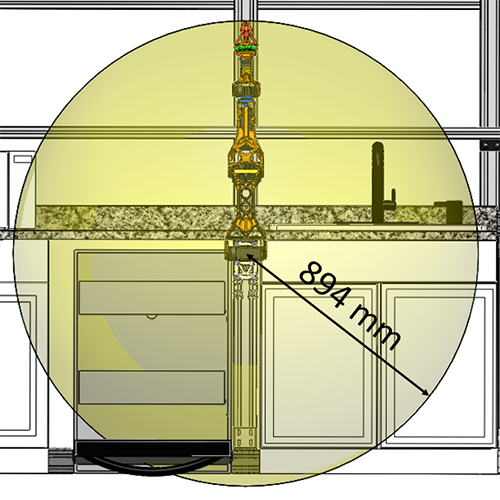}
        \caption{Kitchen (front)}
        \label{fig:kitchen_front}
    \end{subfigure}
    \caption{Visualization of workspaces. Most rectangular dinner tables are 914.4mm (36") wide. The standard dishwasher and sink dimensions (LxWxH) are typically 609.6x609.6x889 mm and 762x558.8x254 mm, respectively.}
    \label{fig:workspace}
\end{figure}

\subsection{Design Considerations}\label{sec:2a}
From the task-level point of view, the location of the robotic arm's base in the target task space is essential to determine the robot's workspace. In a manipulation task with one robotic arm, controlling the robot means moving the relative position and orientation of its end-effector frame $\{e\}$, based on the robot's base frame $\{b\}$ in a world frame $\{w\}$. When there are $n$ objects in the world, we can represent a set of $n$ frames as  $O = \{\{o_1\}, \{o_2\}, ..., \{o_n\}\}$, where $\{o_k\}$ is the local frame attached the $n_{th}$ object $(1\leq k \leq n)$. In most cases, the relation between $\{w\}$ and $\{b\}$, $T_{wb}$, is fixed and given, and the information related to $\{o_k\}$, $T_{wo_k}$, can be estimated through the perception. From this information, we can get the transformation $T_{bo_k}={T_{wb}}^{-1}T_{wo_k}$ and use it for control and planning, considering environments and human collaborators. If the set of target objects $O$ is bounded within a specific area, the optimal location of the robot's base $\{b\}$ to handle all the objects can be found for the target task. For example, while manually washing dishes in a kitchen, the dishes would stay in the kitchen sink. Or, when using a dishwasher, the positions of the kitchen items would be bounded within a certain area including a sink and a dishwasher. Given a robotic manipulator and kitchen information, one can find the proper position and orientation of the robot's base $\{b\}$ for this task.

\begin{figure}
\centering
\includegraphics[width=0.75\columnwidth]{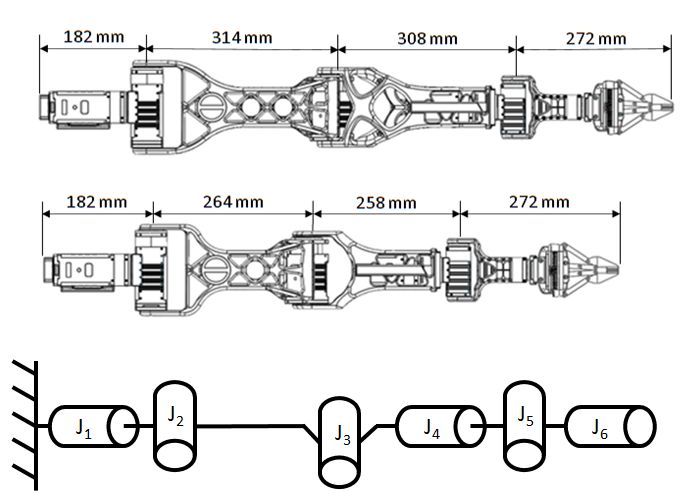}
\caption{PAPRAS long version (top) and short version (middle) in front view, and joint configuration (bottom).}
\label{fig:configuration} 
\end{figure}

\begin{table}
\centering
\begin{tabular}{|c c|} 
 \hline
 Items  & Specifications\\ [0.5ex] 
 \hline\hline
  Mass & Short Version: $4.771$ kg \\
    & Long Version: $4.894$ kg \\
 \hline
 Payload & Short Version: $3$ kg \\
    & Long Version: $2.5$ kg \\
 \hline
 Range of Motion & Joint 1: $-180^{\circ}\sim 180^{\circ}$ \\ 
  ((): Long version)   & Joint 2: $-115^{\circ}\sim 115^{\circ}$  \\ 
   & Joint 3: $-135^{\circ} (-45^{\circ})\sim 135^{\circ}$ \\   
    & Joint 4: $-158^{\circ}\sim 158^{\circ}$ \\ 
    & Joint 5: $-90^{\circ}\sim 90^{\circ}$ \\ 
    & Joint 6: $-180^{\circ}\sim 180^{\circ}$ \\ 
 \hline
Max speed & Joint 1, Joint 2: $174^{\circ}/$s \\ 
(continuous)    & Joint 3, Joint 4: $175.2^{\circ}/$s \\ 
   & Joint 5, Joint 6: $175.2^{\circ}/$s \\  
 \hline
 Max Torque & Joint 1, Joint 2: $44.7$ Nm \\ 
(continuous)  & Joint 3, Joint 4: $25.3$ Nm \\ 
   & Joint 5, Joint 6: $5.1$ Nm \\  
 \hline
\end{tabular}
\captionof{table}{Specifications}
\label{table:specs}
\end{table}

The dimensions of PAPRAS were determined considering the target task spaces, which are the tabletop and dishwashing areas. As shown in Fig. \ref{fig:workspace}, the length of the arm's moving part needs to be around 800-900~mm long for these tasks, and its DoF (degree of freedom) needs to be at least 6. To meet these requirements, PAPRAS was developed based on a 6-DoF open-source robotic arm, OpenMANIPULATOR-P~\cite{OpenManipulator}. We modified its design and developed two variants of PAPRAS in Fig. \ref{fig:configuration}, one version being 100 mm longer than the other. 

We define the link between Joint $n$ and Joint $(n+1)$ as Link $n$, and the arm can be broken down into five links and six joints. The placement and orientation of each joint are shown in Fig.~\ref{fig:configuration}. Joints 2, 3, and 5 are pitch joints and their rotation axes are parallel. However, they are not in the same plane because the position and orientation of Joint 3 have offsets. The offsets make PAPRAS fully foldable around Joint 3. Since the shorter arm has the same configuration of the OpenMANIPULATOR-P, detailed information can be found in~\cite{OpenManipulator}. For all the joints, Dynamixel-P series motors are used. Table~\ref{table:specs} shows the specs of all the joints including the range of motion, max speed, and torque information.  An arm has a gripper with a camera at the output of Joint 6. The gripper and camera are commercialized products, RH-P12-RN from ROBOTIS and RealSense D435 from Intel, respectively.

\subsection{Linkage Design}
\begin{figure}
\centering
    \begin{subfigure}{1\columnwidth}
        \centering
    \includegraphics[width = .8\columnwidth]{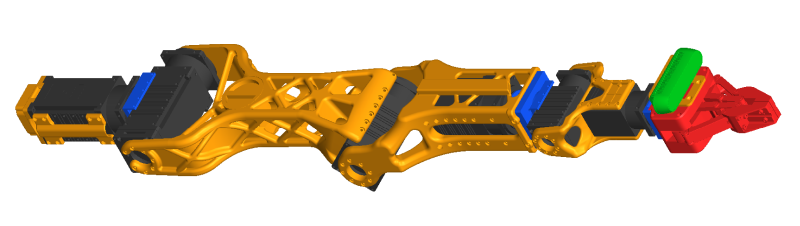}
    \caption{PAPRAS colored by parts. Motors in black, 3D-printed parts in orange, connectors (commercialized parts) in blue, a gripper in red, and a camera in green.}
    \label{fig:breakdown}        
    \end{subfigure}
    \newline
    \newline
    \begin{subfigure}{1\columnwidth}
        \centering
    \includegraphics[width = .9\columnwidth]{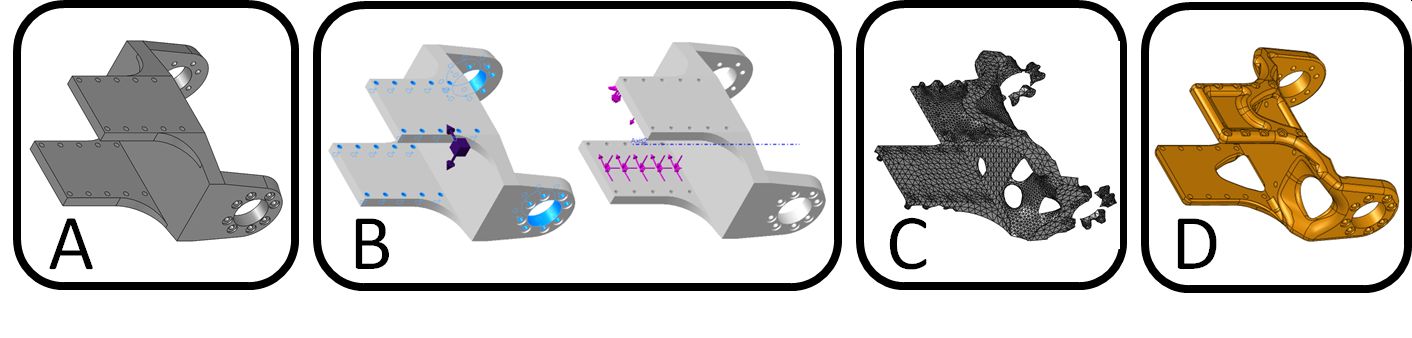}
    \caption{An example (Link 5) of the linkage weight reduction by structure optimization. A) initial design, B) applied constraints, C) topology optimization, D) the resulting part.}
    \label{fig:opt}        
    \end{subfigure}
\caption{PAPRAS parts break down and linkage weight reduction process.}
\label{fig:sub} 
\end{figure}

For the linkage design, we aimed to reduce the overall weight of the links. This was done to reduce the moments of inertia about the joints and thus decrease the torques required to perform motions. As shown in Fig.~\ref{fig:breakdown}, Links 1 and 4 are commercialized parts, and Links 2, 3, and 5 are 3D-printed parts. All 3D-printed components in PAPRAS were made using two FDM printers (Flashforge Creator 3, Raise 3D E2), and were printed using standard PLA filament. The use of 3D-printed PLA not only reduced the weight of the components but also allowed other features to be incorporated into the design.

In order to reduce the weight of the new 3D-printed links, we used the topology optimization tool in Solidworks. This feature allowed us to selectively remove sections from each link while simultaneously maximizing the stiffness-to-weight ratio. To use this feature, the implementation of some constraints was needed to get the desired output. These constraints ensured that the parts had the desired symmetry and did not remove desired geometric features such as faces/holes which connected with the motors. Fig.~\ref{fig:opt} shows the process to get the optimized design for Link 5. We used the same process for Link 2 and Link 3 to reduce the weight while keeping the target stiffness we set. After each design, loads were simulated using the FEA tool in Solidworks to test the design.

\begin{table}
\centering
\begin{tabular}{|c c c c|}
 \hline
 Component & MANIPULATOR-P & Short Version & Long Version\\ [0.5ex] 
 \hline\hline
 Link 2 & 644.0 g & 231.9 g & 277.4 g \\ 
 \hline
 Link 3 & 496.0 g & 211.6 g & 289.1 g \\
 \hline
 Link 5 & 114.0 g & 35.7 g & 35.7 g \\
 \hline
\end{tabular}
\captionof{table}{ Link weight comparison}
\label{table:weight}
\end{table}

In the long version of PAPRAS, Links 2 and 3 are both $50$~mm longer than those in a short version. In Fig.~\ref{fig:configuration}, the long version (top) has a more complicated structure than the short version (middle). Link 3 of the long version has a reinforced structure near Joint 3 to meet the stiffness constraints. This makes Link 3 slightly heavier, and Joint 3 has a smaller range of motion than Joint 3 in a long version. 



\begin{figure}
    \centering
    \begin{subfigure}{.23\textwidth}
        \centering
    \includegraphics[width = 0.75\columnwidth]{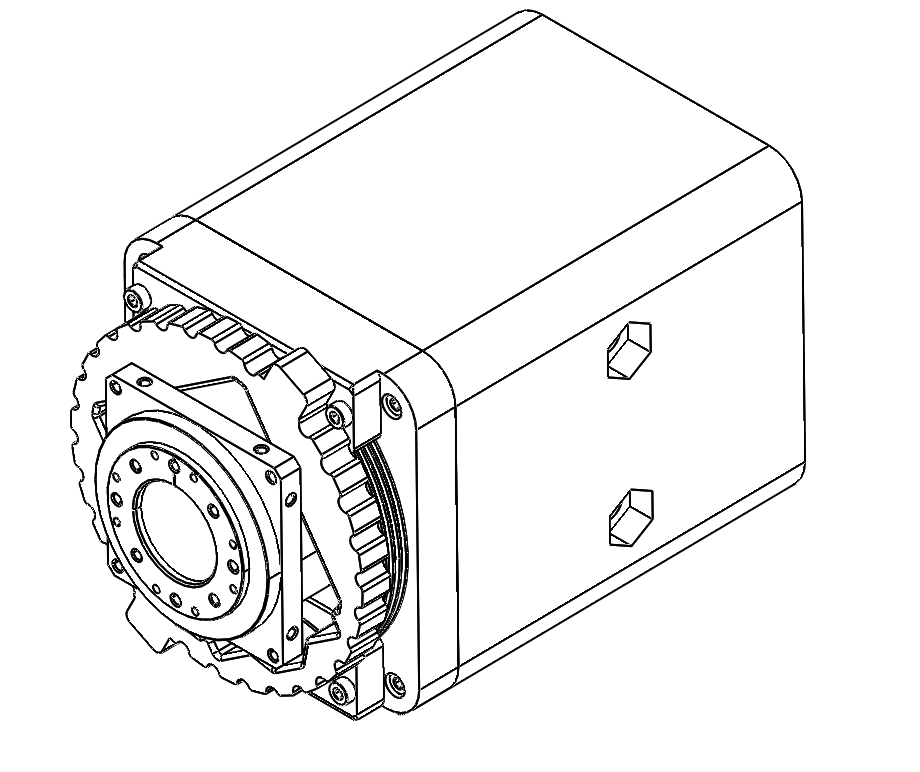}
    \caption{Mount with Joint 1.}
    \label{fig:mountlock}        
    \end{subfigure}
    \begin{subfigure}{.23\textwidth}
        \centering
    \includegraphics[width = 0.75\columnwidth]{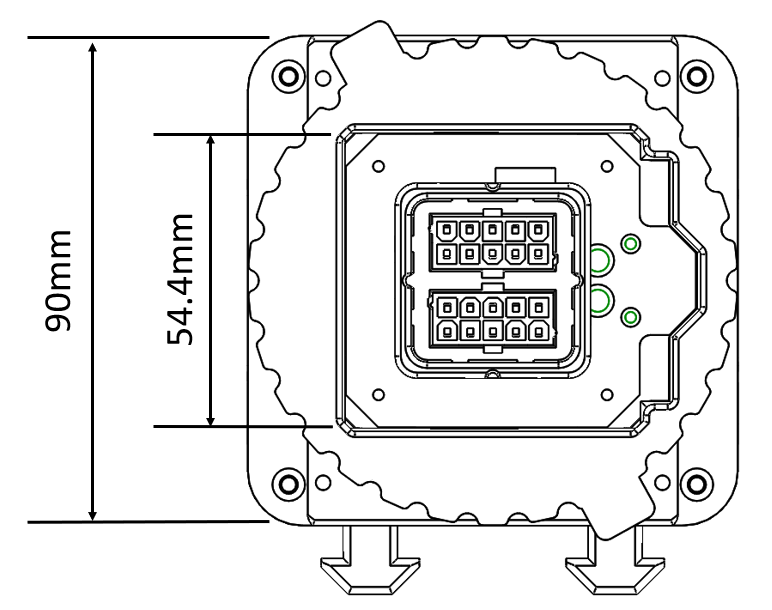}
    \caption{Mount dimensions.}
    \label{fig:mountdimenstion}        
    \end{subfigure}
    \newline
    \newline
    \begin{subfigure}{0.95\columnwidth}
    \centering
    \includegraphics[width = 0.6\columnwidth]{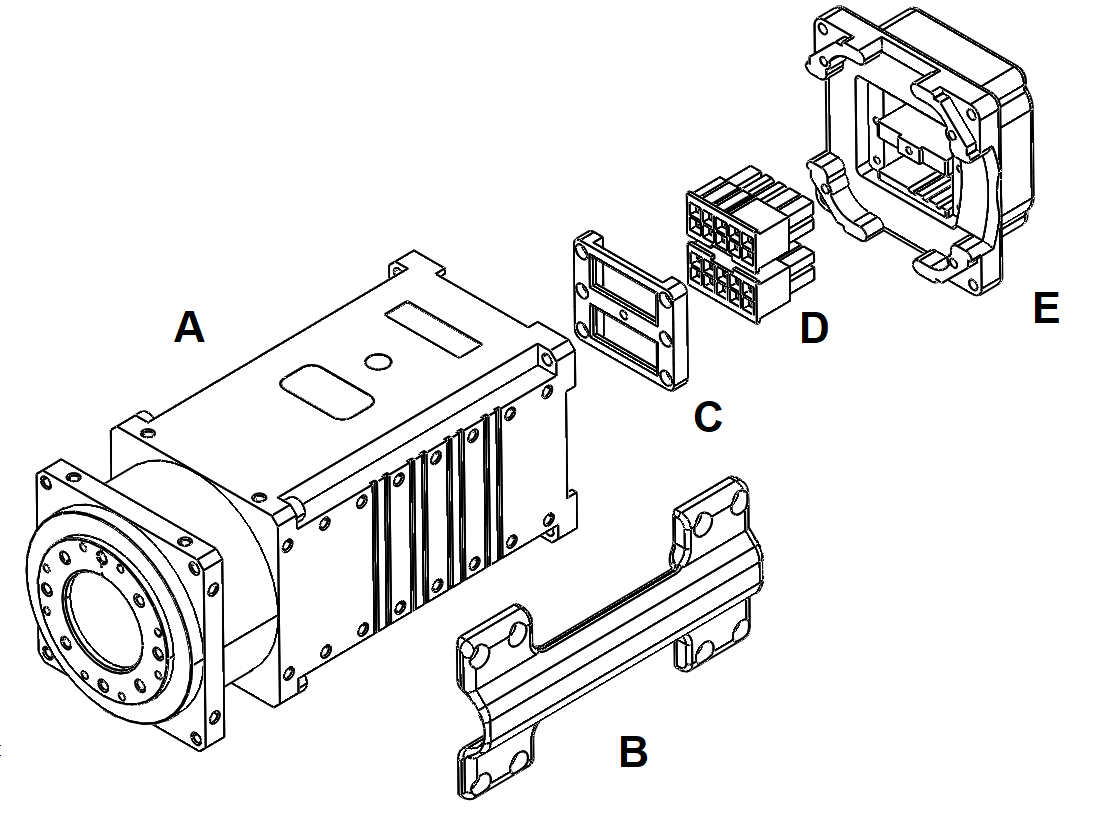}
    \caption{Male section, exploded view. A) Joint 1, B) 3D-printed wire cover, D) two sets of 10-pin Molex connectors (male), C, E). 3D-printed parts.}
    \label{fig:male}        
    \end{subfigure}
    \newline
    \begin{subfigure}{0.95\columnwidth}
    \centering
    \includegraphics[width = 0.6\columnwidth]{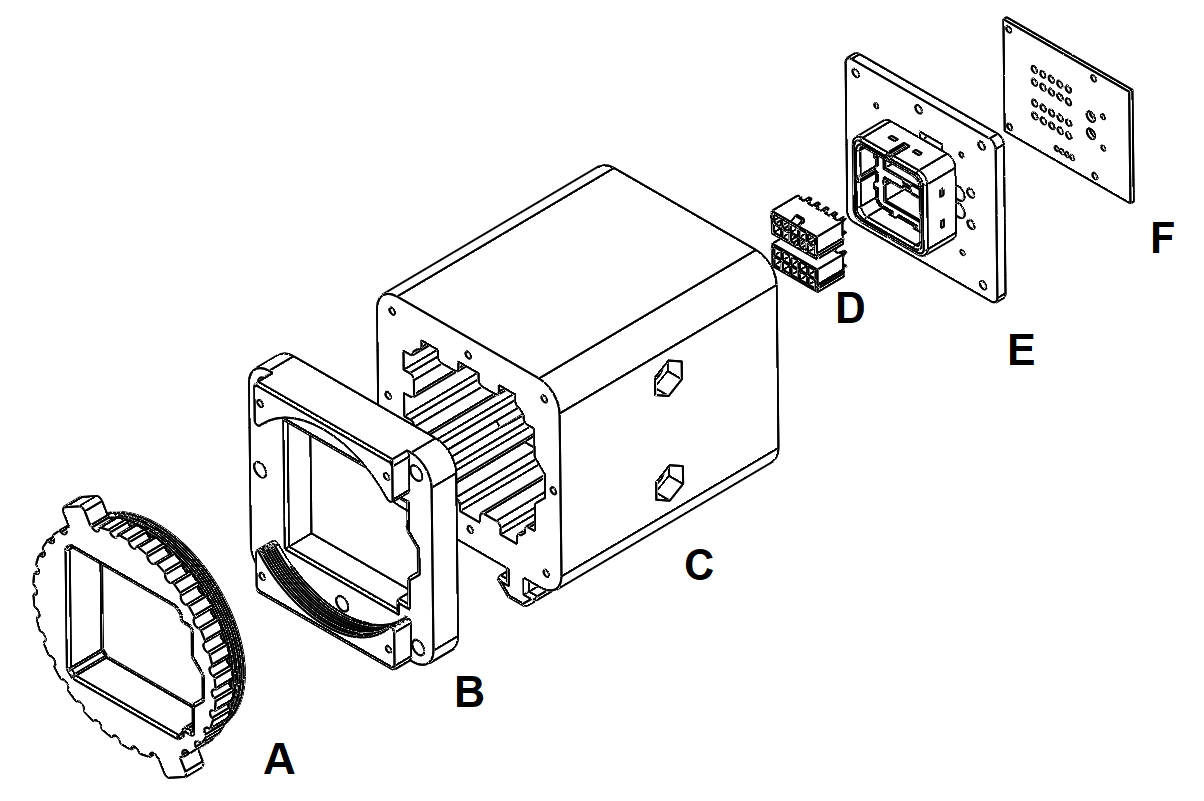}
    \caption{Female section, exploded view. A, B) rotating locking mechanism, C) body of the socket, D) 2 sets of 10-pin Molex connectors (female), E) 3D-printed back plate, F) custom-made PCB.}
    \label{fig:female}        
    \end{subfigure}
\caption{Docking mount.}
\label{fig:dockingmount}
\end{figure}

\subsection{Docking mount}
The docking mount of PAPRAS is composed of male and female sections. The male section in Fig.~\ref{fig:male} is composed of Joint 1 (a motor), along with two sets of 10-pin Molex connectors and three 3D-printed components. The female section in Fig.~\ref{fig:female} is composed of four 3D-printed components which allow for a locking mechanism to keep the motor in place when in use. The mechanism works by having a threaded component on the female section; when this section is screwed in it presses itself onto the motor, securing it in place as shown in Fig.~\ref{fig:mountlock}. When it is unlocked as shown in Fig.~\ref{fig:mountdimenstion}, the arm can be unplugged easily by pulling it out.

Four different types of these female sections were used in this paper: one for the table demo that is made to be orientated vertically, one for the dual-arm and cage demos that is made to orientated horizontally, with a modified horizontal version used for the kitchen demo, and one for the quadruped demo that is made to be orientated at an angle ($45^{\circ}$). The horizontally orientated one is shown in Fig.~\ref{fig:dockingmount}, and all the mounts have the exact locking mechanism.

\begin{figure}
\centering
    \begin{subfigure}{1\columnwidth}
    \centering
    \includegraphics[width = .9\columnwidth]{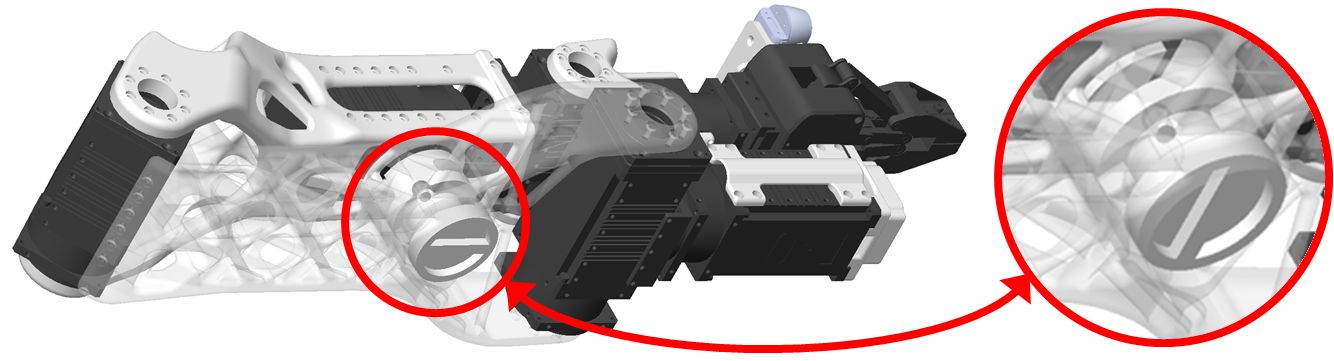}
    \caption{Joint 3 in unlocked position (left) and locked position (right).}
    \label{fig:lock&unlock}        
    \end{subfigure}
    \newline
    \newline
    \begin{subfigure}{.9\columnwidth}
        \centering
    \includegraphics[width = .7\columnwidth]{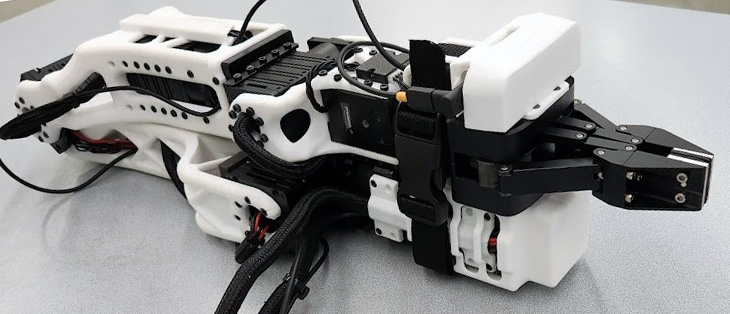}
    \caption{PAPRAS with covers attached.}
    \label{fig:cover} 
    \end{subfigure}
\caption{Portability features of PAPRAS.}
\label{fig:portability} 
\end{figure}

\subsection{Features for portability}
One design goal of PAPRAS is portability. To achieve this, the weight reduction in the previous section is important. At the same time, an easy-to-carry design should be considered. There are two key features for this in PAPRAS. 

\subsubsection{Locking mechanism} PAPRAS is designed with a locking mechanism, which holds the arm in a closed position when it is not in use. The mechanism is based on a rotating cam in Link 2 which hooks onto a protrusion from Link 3. When the mechanism is in the unlocked position it is held in place with magnets. The arm can be seen in both locked and unlocked positions in Fig.~\ref{fig:lock&unlock}.

\subsubsection{Covers for both ends} The gripper/camera module and Joint 1 with the connectors are located at each end of PAPRAS. When PAPRAS is fully folded around Joint 3 and locked with the locking mechanism, these parts lie one upon another as shown in Fig.~\ref{fig:lock&unlock}. Both parts are relatively feeble to impact and need to be fixed in order to not move around Joints 2 and 5. To prevent damage and lock their positions, a cover system was created. Two parts of the cover system were 3D-printed and a cloth strap was used to connect the two parts. Fig.~\ref{fig:cover} shows a folded PAPRAS (short version) with the camera and socket covers attached.



\section{Electrical Hardware}

The electrical hardware for PAPRAS consists of three parts: the wiring harness on each arm, the connection on each mount, and the control box for each application.

\subsection{Arm Wiring Harness}

The motors are all connected to an RS-485 serial communication bus, which is daisy chained up the arm to each motor and the gripper. The Joint 1-4 motors each require external power, while the Joint 5-6 motors and the gripper are powered by the RS-485 voltage. There is also a USB-C cable that runs the length of a PAPRAS arm to connect to a USB camera mounted on the gripper. All these wires are routed to two 10-pin male connectors located at the base of the arm, which mate with connectors in the mounts.

\subsection{Mount Connection}

Each mount has a custom PCB (F in Fig.~\ref{fig:female}) which consists of the connections responsible for distributing power to the arm and transferring data to and from the arm. The PCB has two 10-pin female connectors that mate with the matching set on each arm. Of the twenty pins, four are used for the RS-485 serial communication with the motors. Eight more pins are used for the external power connections required by the Joint 1-4 motors. The final eight pins are used for the USB 3.1 connection between the computer in the control box and the USB camera on the gripper. The PCB also contains connectors that lead back to the control box – an XT connector for power, a 4-pin Molex connector for serial communication, and a USB 3.1 Type A receptacle.

\begin{figure}
    \centering
    \begin{subfigure}{0.40\columnwidth}
        \centering
        \includegraphics[height=4cm]{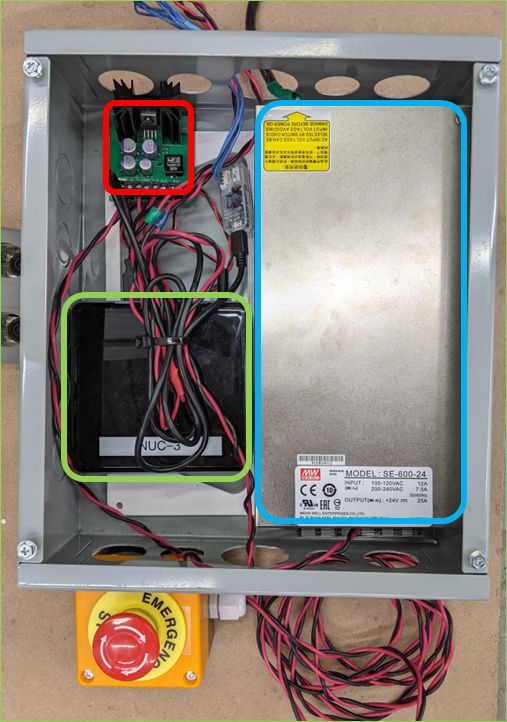}
        \caption{Control box.}
        \label{fig:fixed}
    \end{subfigure}
    \begin{subfigure}{0.40\columnwidth}
        \centering
        \includegraphics[height=4cm]{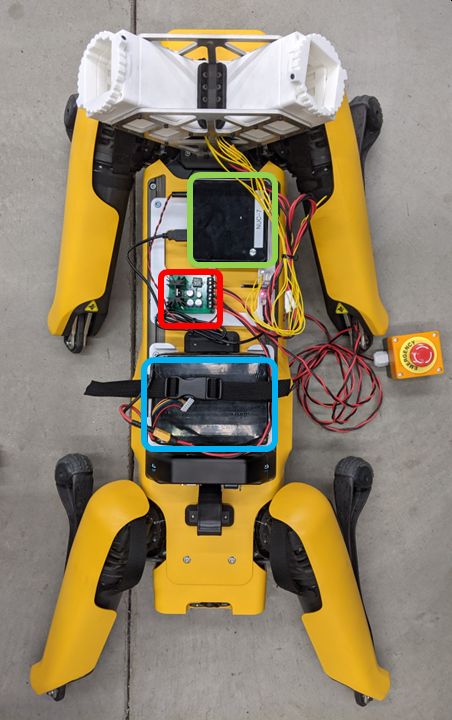}
        \caption{Spot application.}
        \label{fig:spot}
    \end{subfigure}
    \caption{Control box for a fixed PAPRAS application (a) and setup for an untethered mobile application (b). The power source (blue), controlling computer (green), and custom PCB (red) are highlighted.}
    \label{fig:control_setup}
\end{figure}

\subsection{Control Box}

Each application has a single control box regardless of how many mounts and arms are involved. The control box contains a power supply, a controlling computer, an RS-485 adapter (U2D2), an emergency stop, and another custom PCB. The power supply connects to AC mains and outputs 24V DC power, which is routed through the emergency stop switch before running to the mount(s) to power the arm motors. The power supply also feeds a buck regulator on the PCB (independent from the emergency stop) which outputs 19V to power the computer. This setup allows for the e-stop to safely cut power to the arms without dropping power to the controlling computer. The computer connects to the U2D2 over USB, which converts the USB signal into the RS-485 signal required to interface with the motors. The serial nature of the communication protocol allows multiple mounts to connect in parallel with a single U2D2. The computer also can connect directly to the USB receptacle on the mount(s) to communicate with the camera on the gripper.

For mobile applications, the power supply is replaced with a 6S LiPo battery. The battery power is used for both the arm motors and the controlling computer. The emergency stop switch and the communication setup are identical to fixed applications. This change keeps the mobile platform untethered and free to move to any location. Fig.~\ref{fig:control_setup} shows a control box for a fixed application as well as the setup for a mobile application using Boston Dynamic’s Spot.




\section{Software Architecture}
\begin{figure}
    \centering
    \includegraphics[width=\columnwidth]{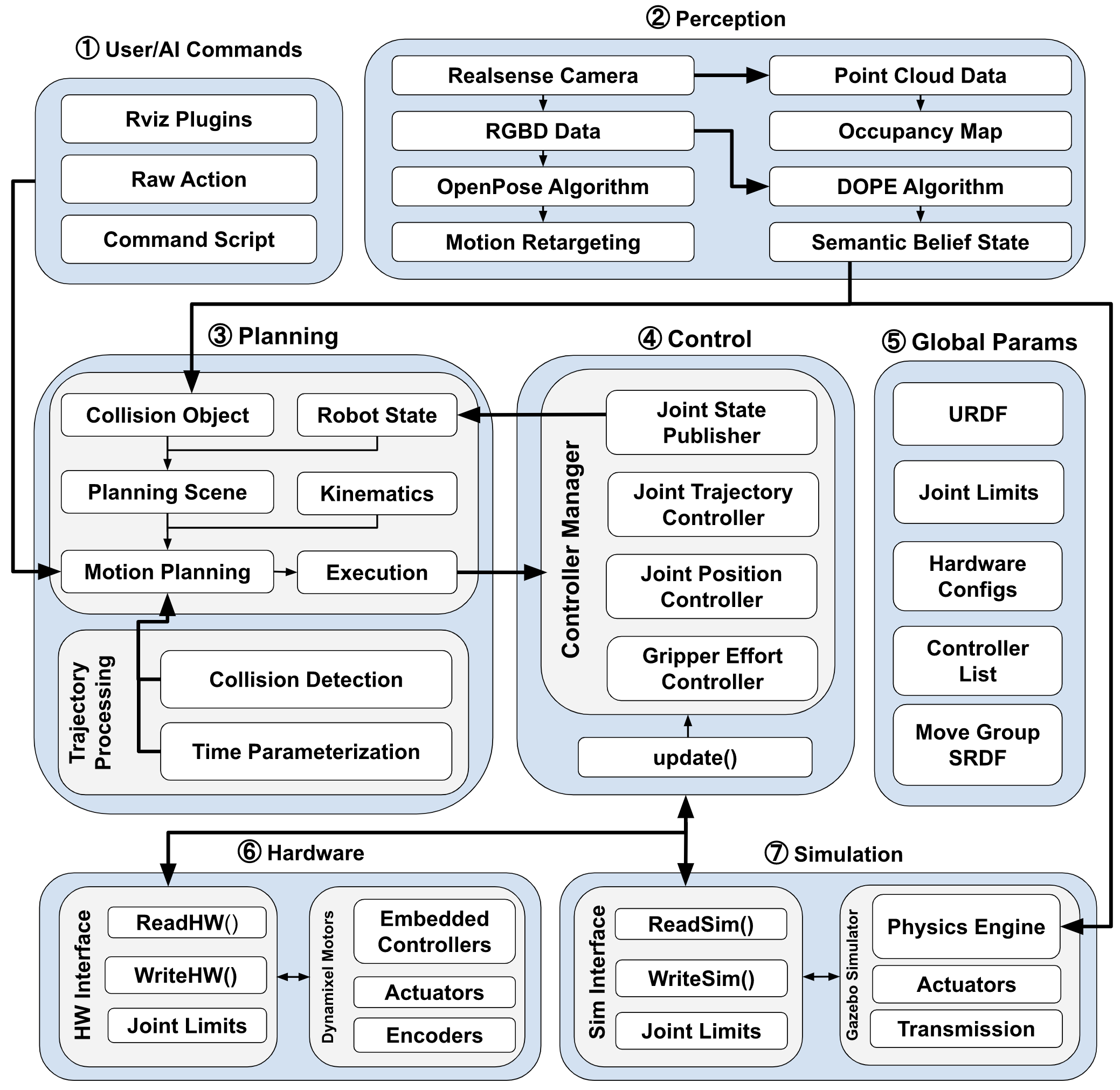}
    \caption{Overview of software architecture.}
    \label{fig:software_arch}
\end{figure}

PAPRAS's software architecture is built on an open-source framework and optimized for low-latency multiagent-based distributed manipulator control. Developed in C++, the architecture runs on ROS Noetic with the Ubuntu 20.04 Linux distribution. Low-level programming for motor sub-routine calls employs the Dynamixel SDK. High-level programming of the robot, such as the planning and perception pipelines, is accomplished through the incorporation of various ROS packages. A breakdown of the complete flow of the software pipeline is shown in  Fig.~\ref{fig:software_arch}, including user operation, perception, planning, control, simulation, and hardware. 

\subsection{Global ROS Parameter Server}

The Global ROS Parameter Server, shown in \textcircled{5} of Fig.~\ref{fig:software_arch}, is a core component of the software architecture that stores data and settings that are shared across all active nodes. It includes the URDF and SRDF files, a list of controllers, hardware configurations, and joint limits. This data is used to initialize PAPRAS demo applications and is critical for the proper functioning of the robot. The URDF defines the environment, components and geometric structure of the robot arms, while the SRDF assigns planning groups, poses and collision checking information. Joint groups consist of the chain of links from the first link to the end-effector link, and can also be broken into sub-groups. Default poses like ‘init’, ‘rest’, ‘open’ and ‘close’ are assigned for testing and safe positioning, and the SRDF can tweak the collision checking information to reduce trajectory validation time.

\subsection{User-based Operation and Monitoring}
As shown in \textcircled{1} of Fig.~\ref{fig:software_arch}, users can operate the system in three ways: using a 3D marker in Rviz, with joystick or keyboard inputs, or by running a command script which contains joint angles or cartesian-based end-effector poses for each step of a given task. The user can command robots to reach desired configurations either by sending a command to the planning pipeline (\ref{planning_pipeline}) to generate a valid trajectory, or alternatively by sending the command directly to the controller manager (\ref{controller_manager}) to move directly to the goal position. The process is monitored with the help of RQT and Rviz plugins which provide debugging tools during development and validation testing.

\subsection{Perception Pipeline}
\label{perception_pipeline}

\begin{figure}
    \centering
    \includegraphics[width=0.8\columnwidth]{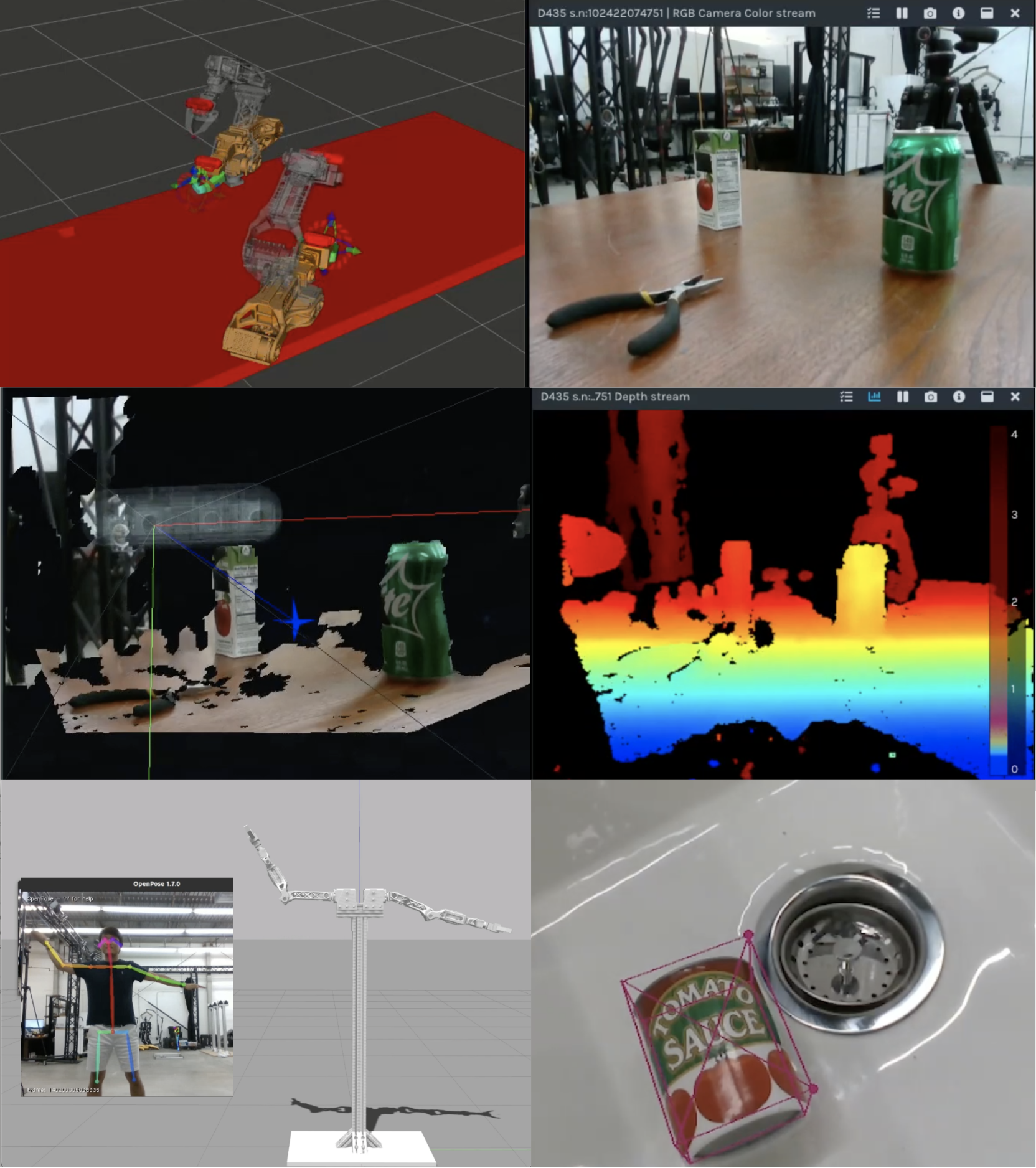}
    \caption{Visual of planned robot motion (top left), RGB (top right), point cloud (mid left), and depth data (mid right), motion retargeting (bot left), object pose estimation (bot right).}
    \label{fig:rgb_pc}
\end{figure}

The perception pipeline, shown in \textcircled{2} of Fig.~\ref{fig:software_arch}, is responsible for obtaining data from the camera on the gripper of each arm, with a sampling rate of 15Hz. Fig.~\ref{fig:rgb_pc} demonstrates the RGBD image and point cloud data of a table environment. This data is used in various perception algorithms to give the robot increased autonomy and decision-making capabilities. For example, the DOPE (Deep Object Pose Estimation) algorithm can recognize objects and estimate their 6 DoF pose \cite{dope2018}. This pose is recorded as a transformation from the camera frame to the object frame, $T_{co}$. This data is used by the pipeline to create an  occupancy map, which helps the robot to build a semantic belief state of its surroundings. Additionally, OpenPose human skeleton tracking can be used to extract corresponding joints from the 2D image \cite{openpose2019}. This information can then be retargeted to the robot arm. This technology is demonstrated as a demo application in Section~\ref{applications}. All of these perception algorithms provide the robot with increased control over its environment.


\subsection{Planning Pipeline}
\label{planning_pipeline}

The planning pipeline, shown in \textcircled{3} of Fig.~\ref{fig:software_arch}, uses MoveIt to manage the planning scene and trajectory generation. The planning scene is used to represent the environment around the robot arms and their current state by processing the joint states of each motor in the arm. Objects of interest can be added to the planning scene manually or by using a 2D or 3D camera to detect them. By using the move group interface, requests can be made to the planning scene to move to a certain pose defined in either joint space, cartesian space, or by an object to interact with. Goal poses must be defined in the base frame of the arm. To calculate the goal pose for interacting with an object, the transform $T_{co}$ found by the perception pipeline must be transformed into the correct frame with the equation $T_{bo} = T_{bc} T_{co}$, where $T_{bo}$ is the goal pose for the arm to interact with the object, and $T_{bc}$ is the pose of the camera in the base frame of the arm, calculated through the forward kinematics defined by the URDF and current joint angles. In order to determine the correct joint angles from a given cartesian space goal, the TRAC-IK numerical inverse kinematics solver \cite{Track2015} is then invoked. The goal state is then defined in joint space and a trajectory to reach it from the current state is found by running OMPL's RRT-Connect planner \cite{sucan2012the-open-motion-planning-library, 844730}, taking into account any constraints that have been specified. Collision checking is performed using FCL \cite{6225337} at each time step of the trajectory to ensure that the returned trajectory is safe. Once the trajectory has been generated, it is sent to the controller manager (\ref{controller_manager}).


\subsection{Controller Manager}
\label{controller_manager}
The controller manager, shown in \textcircled{4} of Fig.~\ref{fig:software_arch}, provides a real-time compatible control loop that runs a rate of 8~ms, and the infrastructure to load, unload, start, and stop controllers. It enables controllers to access joint state information and execute commands from a single interface. The manager is able to read joint states from the hardware and send commands to it. The joint position controller is utilized to send position commands to the arm's six joints. The joint trajectory controller is used for planned trajectory messages, which is done with the help of its follow joint trajectory action server. This action server allows for tracking of trajectory execution, passing goals to the controller, and reporting success when done. The trajectory may also be constrained or aborted if the constraints are broken. Lastly, an effort controller is used for the grippers because current-based position control is needed at the embedded level control.

\subsection{Hardware and Simulation Interfaces}
\label{hw_sim_interface}
The hardware interface, shown in \textcircled{6} of Fig.~\ref{fig:software_arch}, uses the Dynamixel SDK to communicate with the motor controllers through U2D2. Before communication is established, the hardware configuration file is used to relate the motor IDs to the arm configuration. This enables the hardware interface to send commands to the correct actuators and access their parameters. During the execution of a task, the controller manager sends velocity, position, and torque commands to the motors. The Dynamixel SDK then applies these commands to the motors to move them to the desired setpoints and enforce joint limits. To ensure joint limits are not exceeded, the controller manager continuously monitors the encoder positions of the motors, and if the setpoint is outside the joint limits, the motor is commanded to stop.

The simulation environment Gazebo 11 is used to simulate multiple manipulators in complex scenarios. It is loaded using environment-specific URDF files, which contain mass, inertia, and joint configuration information. Gazebo is bridged with ROS using the gazebo\_ros\_lib plugin, allowing the controller manager to interface with the motor input and sensor output. The simulator's physics engine is used to validate the goal action by evaluating the plan. The state of the virtual environment can be synchronized with the real world by updating the simulation through the perception pipeline. As shown in the data flow of \textcircled{7} in Fig.~\ref{fig:software_arch}, the robot can use the updated belief state to spawn 3D models in simulation and planning scenes. 



\begin{figure}
\centering
    \begin{subfigure}{0.9\columnwidth}
        \centering
        \includegraphics[width = 0.88\columnwidth]{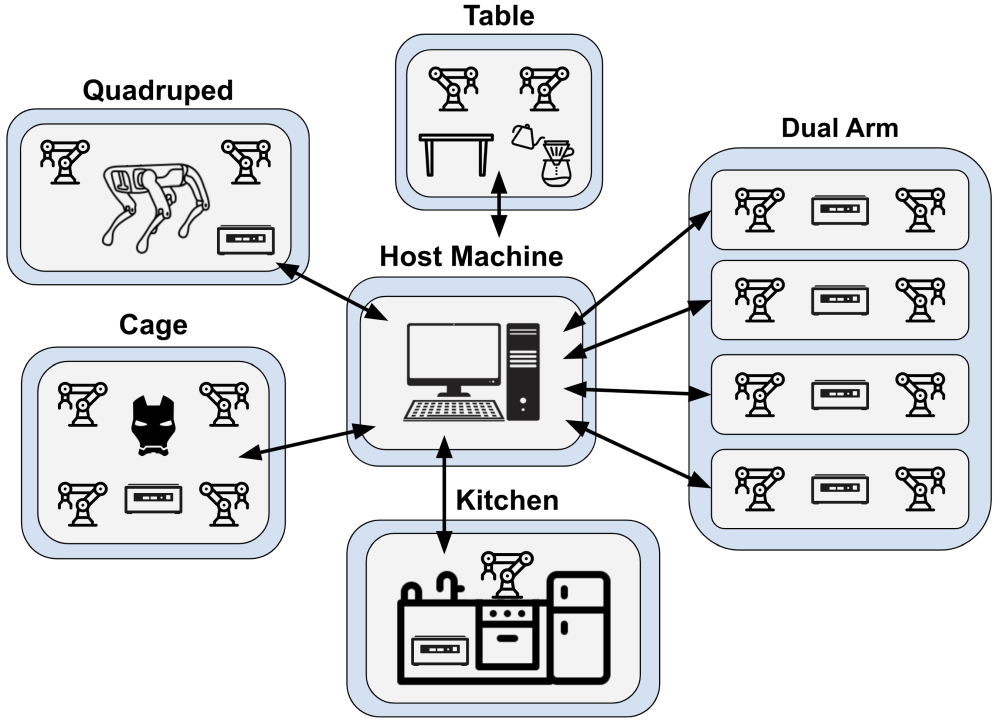}
        \caption{Demo system configurations.}
        \label{fig:demo_sys_config} 
    \end{subfigure}
    \begin{subfigure}{0.9\columnwidth}
        \centering
        \includegraphics[width = 0.8\columnwidth]{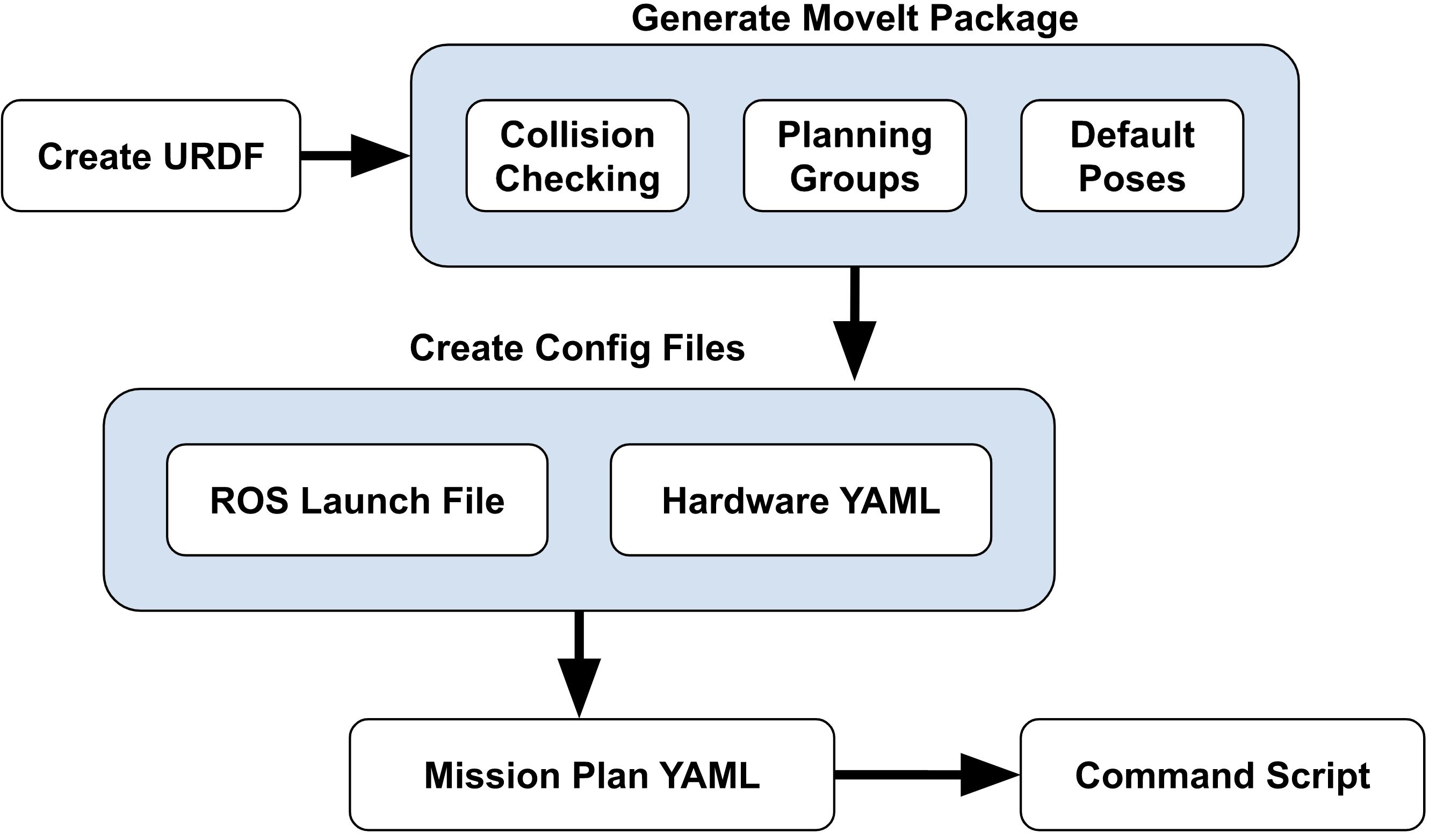}
        \caption{Process to create a new demonstration.}
        \label{fig:new_demo} 
    \end{subfigure}
     \caption{PAPRAS applications.}
     \label{fig:applications}      
\end{figure}

\subsection{Communication between Distributed Computing Units}
\label{comms}
As demonstrated in Fig.~\ref{fig:demo_sys_config}, PAPRAS uses distributed computing capabilities to allow communication between distributed computing units. Using the master ROS node on the main operating computer as the host machine, all other running processes can be configured with their IP and master URI addresses for the connection. The host machine runs the high-level and computationally heavy nodes such as the user or AI-based task commands as well as the perception and planning algorithms. Furthermore, multiple PAPRAS arms can be controlled from each client machine and its movements can be synchronized with other client machines.  The main host machine used is the Lambda Dual GPU Workstation, while the client machines are Intel NUCs composed of an Intel Core i3-10110U processor, 16GB DDR4 RAM, and a 256GB SSD.

\section{Hardware experiments}
\label{experiment}

\subsection{Validations}
\label{validations}
\begin{figure}
\centering
    \begin{subfigure}{0.49\columnwidth}
    \centering
    \includegraphics[height=3.2cm]{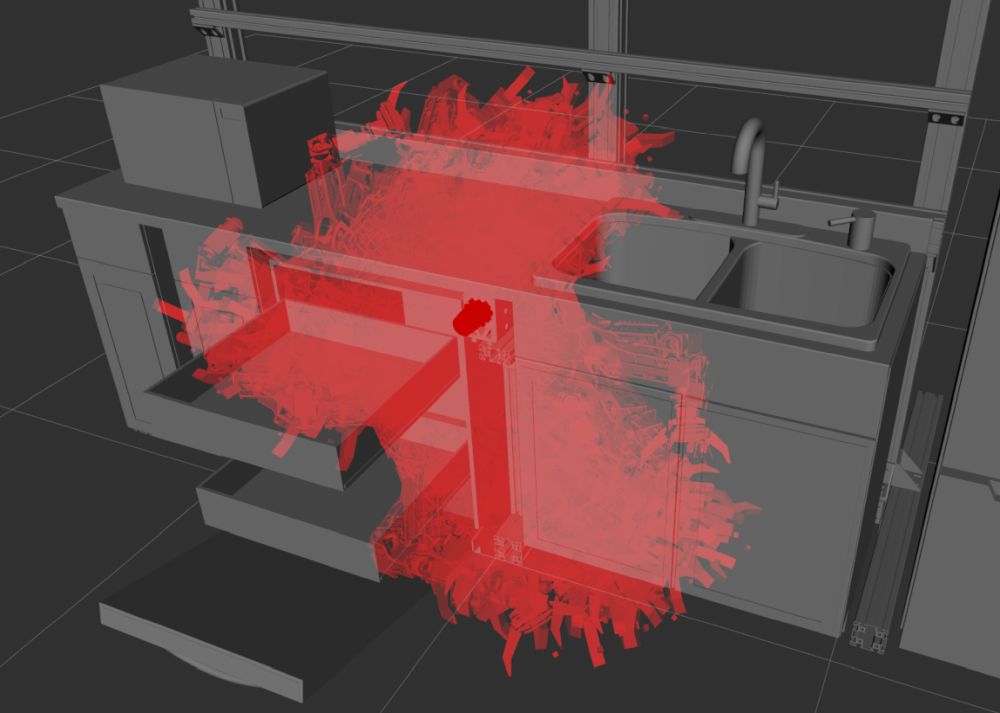}
    \caption{300 poses}
    \label{fig:payload_sim}        
    \end{subfigure}
    \begin{subfigure}{0.49\columnwidth}
        \centering
    \includegraphics[height=3.2cm]{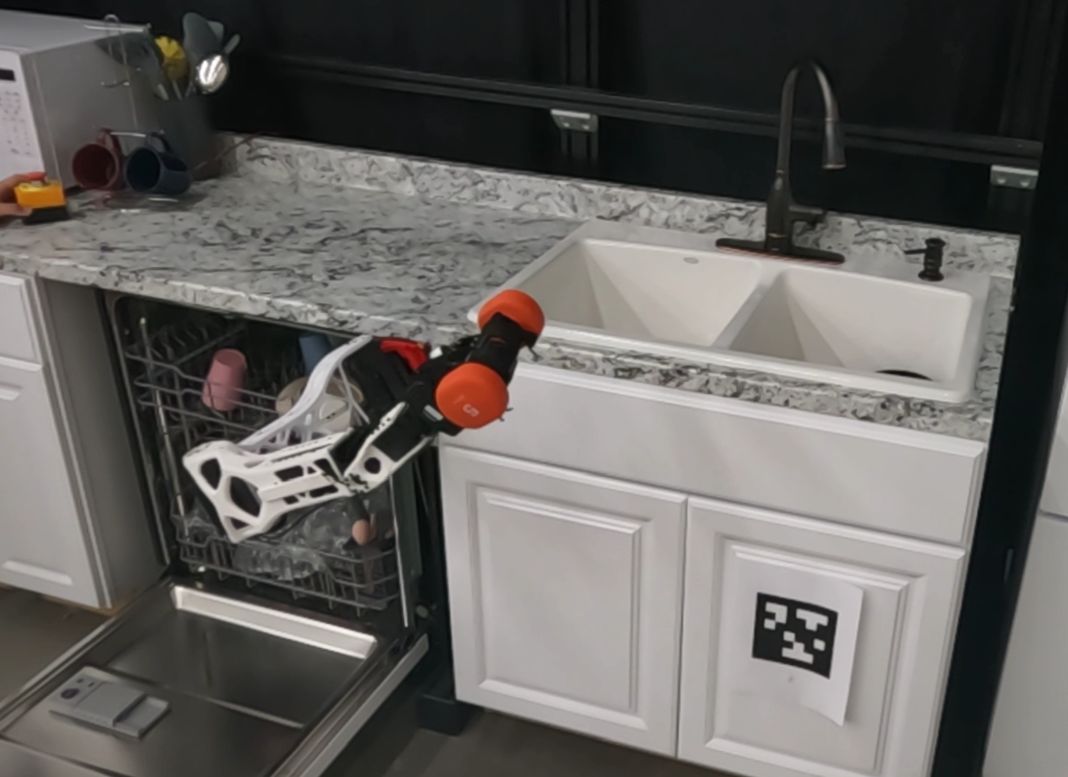}
    \caption{PAPRAS moving 5lbs payload}
    \label{fig:payload_hardware} 
    \end{subfigure}
\caption{Payload check in simulation and experiments.}
\label{fig:validation} 
\end{figure}

Hardware experiments were performed in a kitchen environment to validate the functions and performance of PAPRAS. In Fig.~\ref{fig:validation}, a docking mount (Fig.~\ref{fig:dockingmount}) and a control box (Fig.~\ref{fig:fixed}) were installed between the dishwasher and the kitchen sink. The mount is located right below the kitchen top facing perpendicular to the surface of the dishwasher door and the cabinet door, as planned in Fig. \ref{fig:workspace}. With this mount location, PAPRAS is able to reach the objects in the sink and the bottom drawer of the dishwasher. We used a 2.27 kg (5 lbs) dumbbell for the payload test. First, 300 hundred random poses near the boundary of the PAPRAS workspace were selected in simulation (Fig. \ref{fig:payload_sim}). From the 300 poses, we selected 10 poses in which the motor torques were close to the maximum torques. A motion was planned between the 10 poses while avoiding collisions. PAPRAS was able to track the planned motion in both simulation and hardware experiments. Please refer to the supplementary video to see the hardware experiment.

\subsection{Applications}
\label{applications}



\begin{figure}
    \centering
    \includegraphics[width=.95\columnwidth]{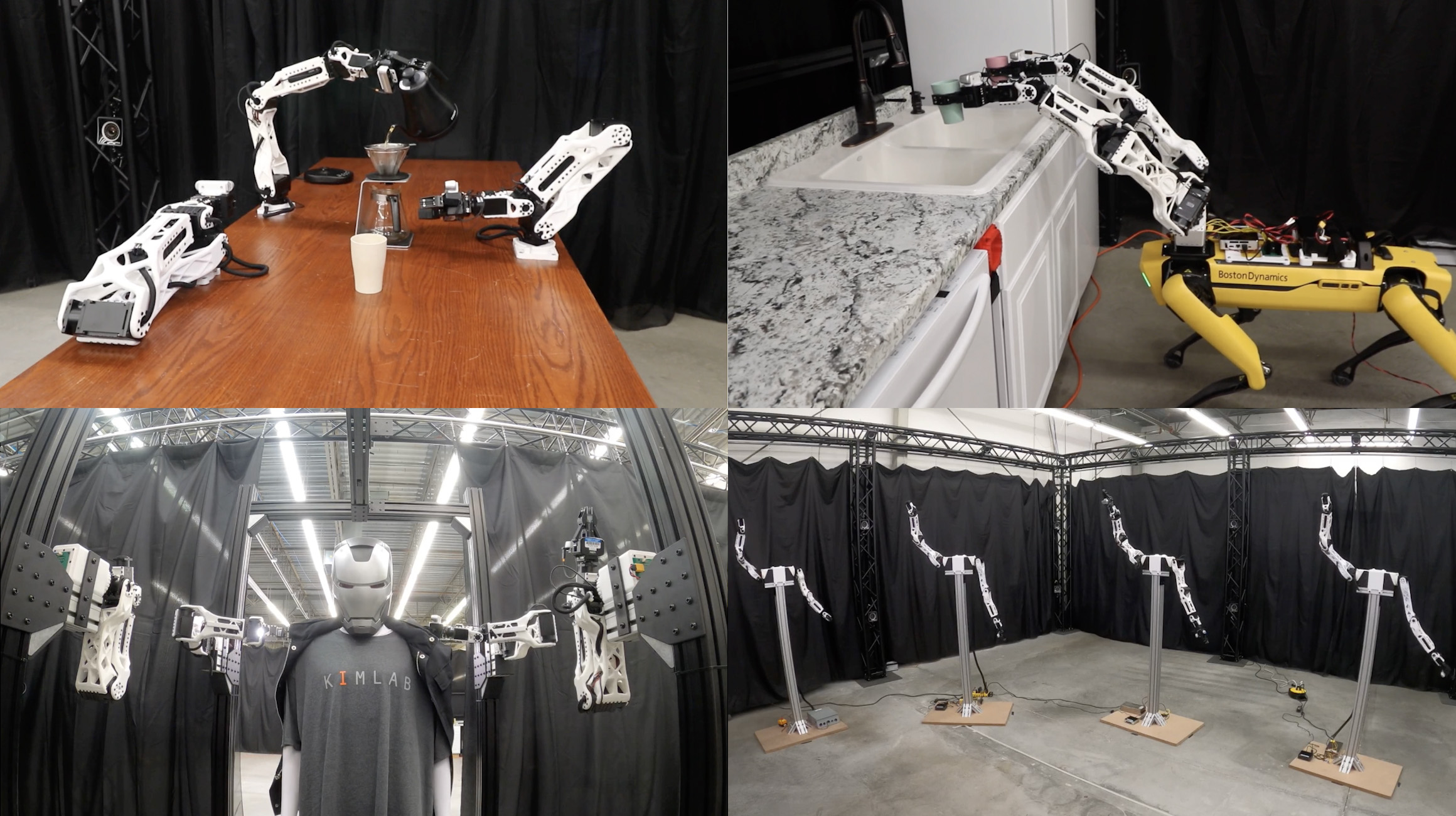}
    \caption{Execution of coffee making, quadruped, cage, and dual arm demos.}
    \label{fig:demos_wide}
\end{figure}

PAPRAS offers a modular framework for quick and efficient implementation of the system in various environments. Fig.~\ref{fig:new_demo} shows how to create a new demonstration. First, a URDF file needs to be created that details the model of the environment and the location(s) and orientation(s) of the PAPRAS mount(s). A MoveIt package needs to be generated with appropriate inter-link collisions disabled and planning groups set up. ROS launch and YAML configuration files need to be created along with a YAML file outlining desired poses and actions. Lastly, a C++ script uses the  Mission Plan to control the arms and gripper. In general, all steps are relatively straightforward and require minimal effort. As a result, PAPRAS allows for the rapid deployment of new demonstrations. Our system demonstrated scenarios such as dishwashing, coffee making, dressing, and dual-arm motions in Fig.~\ref{fig:demos_wide}.

\subsubsection{Sink to Dishwasher}
In the kitchen environment, the goal was to move a set of dirty dishes from the sink to the dishwasher. To do this, we used the same setting as in the payload test. A command script was used to perform the task of localizing and mapping the sink, detecting objects, estimating each item’s pose, choosing the most cost-efficient object to pick,  picking an object from the sink, moving the object to the dishwasher, and placing the object in the dishwasher.

\begin{figure}
    \centering
    \begin{subfigure}{.9\columnwidth}
        \centering
    \includegraphics[width=0.55\columnwidth]{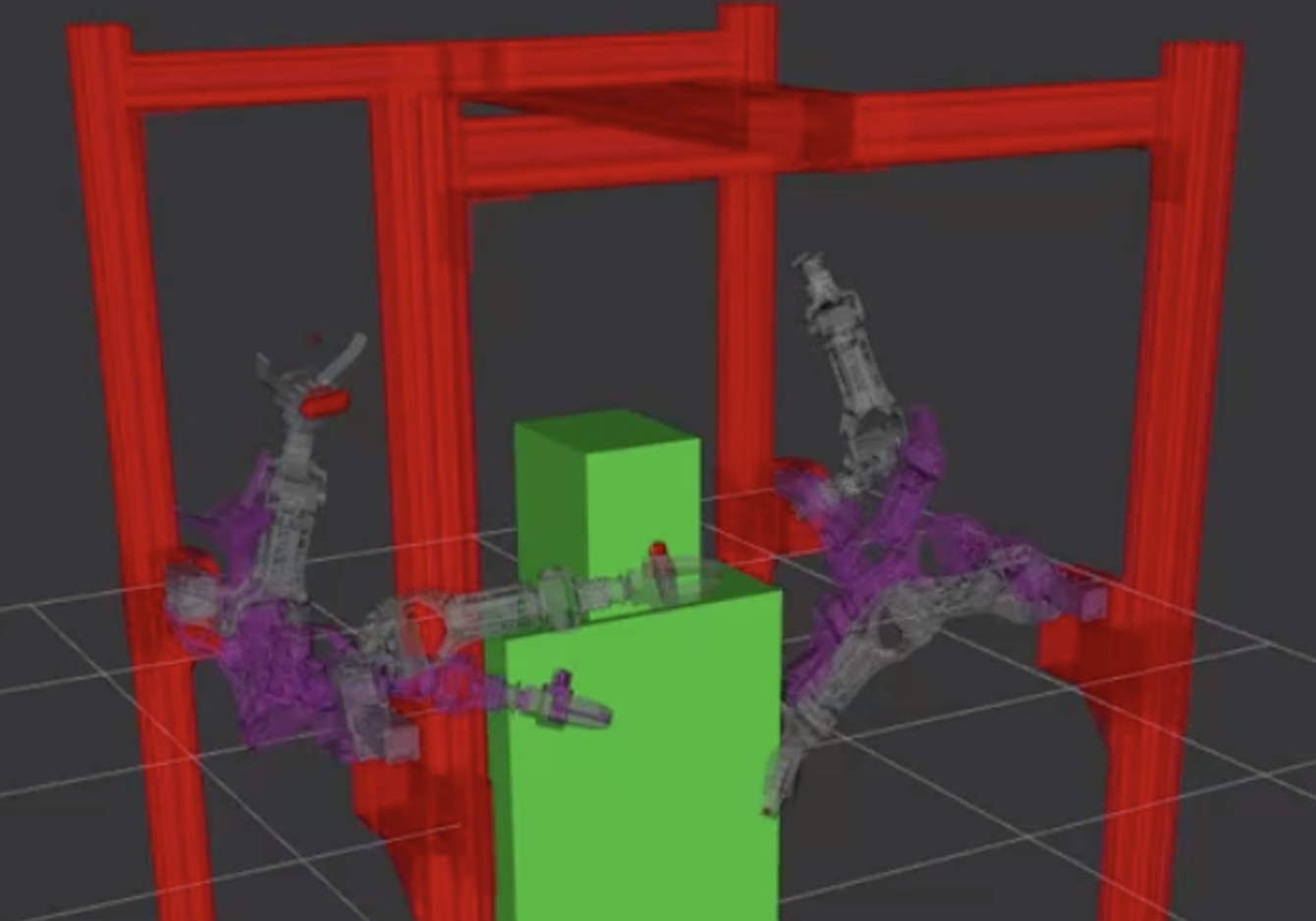}   
    \end{subfigure}    
    \caption{Visualization of cage with collision box.}
    \label{fig:cage}
\end{figure}


\subsubsection{Coffee Making}
For our coffee-making demonstration, we made a dinner table with three arm mounts in Fig. \ref{fig:table_top} and used a cup, an electric kettle, a pour-over coffee cone, and ground coffee. Our command script runs the two stages  of the demo task which consists of pouring periodically from the kettle into the pour-over coffee cone, then serving the coffee into a cup without spilling. 

\subsubsection{Quadruped}
PAPRAS can also be extended to mobile robots, such as Spot. In this application, two arm mounts were attached to the front end of Spot to pick and place objects between the table and sink, where mobility is needed. A command script coordinates arm control operations with mobile navigation while communicating with the Spot SDK.

\subsubsection{Cage}
In the cage demo, four arm mounts were placed on the four vertical beams of the suit-up environment. Here, all four arms were simultaneously controlled by a single client computer to assist the subject with putting on a jacket. The task planning pipeline starts with the back arms holding the end sleeves, bringing the jacket forward for the two front arms, helping the subject step in, and finally bringing down the jacket as the back arms go back to the rest position. This setup enables automated dressing while ensuring safety with real-time multi-agent motion planning. The estimated collision box of the subject seen in Fig.~\ref{fig:cage} was used to ensure that the arms always avoid contact with the user.

\subsubsection{Dual Arm}
The dual arm demo showcases the potential of human-robot interaction and collaboration in real-world scenarios. The platform is built with a single stand, two arm mounts, and a client computer. A camera records the motion of a lead demonstrator, and OpenPose human skeleton tracking is used to directly map the joint angle values from the human to the robot. The mapping is sent as raw action commands to the joint trajectory controller. In addition to the HRI aspect, the dual arm highlights the group communication capabilities between distributed systems. We enable synchronized control of multiple dual arm stands by connecting each client machine to the host machine, as described in \ref{comms}. This communication allows the movements of each robotic arm to be synchronized with other client machines, creating a highly coordinated display.



\section{Conclusion and Future Work}
\label{Conclusion}
This paper presented PAPRAS, a pluggable robotic arm system for home and human-robot collaboration. By analyzing the target task spaces and utilizing 3D printing and structure optimization, we were able to accomplish the lightweight and high-payload arm design. A locking mechanism was embedded in PAPRAS for better portability and a docking mount was implemented for the plug-and-play function. We built the PAPRAS software architecture based on an open-source framework and optimized for low-latency multiagent-based distributed manipulator control. To show PAPRAS's ease of use and efficiency, we developed a process to create new demonstrations. Simulations and hardware experiments were conducted in various demonstrations to validate the hardware and software design. PAPRAS successfully performed sink-to-dishwasher manipulation, coffee making, mobile manipulation on Spot, and suit-up demo.

For future work, we will keep creating more applications in human environments. Furniture, home appliances, and other types of mobile base robots are potential applications of PAPRAS. Furthermore, we are planning to optimize the workspace of PAPRAS based on the 3D data from the target task. As stated in Section~\ref{sec:2a}, collecting spatial data, such as trajectories of the objects in 3D space, is important for the robot to do the task. For this, we will collect the trajectories of the objects and human motions using a motion capture system. From this data, we can get the possible end-effector trajectories for the sequential movements. It will be an interesting design optimization problem to solve various target tasks by optimizing the number of arms, the locations of the mounts, as well as the trajectories of the objects. Computational co-optimization methods, such as \cite{Ha2018}, can be used to find design parameters and motion trajectories for the robotic system at the same time.


\bibliographystyle{IEEEtran}
\bibliography{main}

\end{document}